\documentclass[conference]{IEEEtran}
\IEEEoverridecommandlockouts
\usepackage{cite}
\usepackage{hyperref}
\usepackage{amsmath,amssymb,amsfonts}
\usepackage{graphicx}
\usepackage{algorithm}
\usepackage{algpseudocode}
\usepackage{multirow}
\algrenewcommand{\algorithmicrequire}{\textbf{Input:}}
\algrenewcommand{\algorithmicensure}{\textbf{Output:}}

\usepackage{textcomp}
\usepackage{xcolor}
\usepackage{pgfplots}
\pgfplotsset{compat=1.18}
\usepackage{tikz}
\usepackage{subcaption}
\usepackage{booktabs}
\hypersetup{hidelinks}
\usetikzlibrary{shapes,arrows,positioning,fit,calc}

\def\BibTeX{{\rm B\kern-.05em{\sc i\kern-.025em b}\kern-.08em
    T\kern-.1667em\lower.7ex\hbox{E}\kern-.125emX}}

\usepackage{fancyhdr}
\fancypagestyle{dtsfirstpage}{%
  \fancyhf{}%
  \fancyhead[C]{\small This article is accepted for publication in \textit{IEEE Design, Automation \& Test in Europe} (DATE) 2027}%
}

\begin{document}
\bstctlcite{dts_bib_style}

\title{DTS: Delta-Based Tile Scoring for Efficient Long-Context Sparse Attention\\}

\author{\IEEEauthorblockN{Emre U\u{g}ur\textsuperscript{*}, Jiawen Qi\textsuperscript{*}, Qinyu Chen}
\IEEEauthorblockA{Leiden Institute of Advanced Computer Science (LIACS), Leiden University, The Netherlands}
\thanks{\textsuperscript{*}These authors contributed equally to this work.}
\thanks{Corresponding author: Qinyu Chen (q.chen@liacs.leidenuniv.nl)}
\thanks{The authors acknowledge the Dutch Research Council (NWO) Veni Talent Program (21828) and the LIACS strategic PhD program for funding this work.}
}

\IEEEaftertitletext{\vspace{-1\baselineskip}}
\maketitle
\thispagestyle{dtsfirstpage}

\begin{abstract}
Large language models (LLMs) increasingly need to process long documents and codebases, making long-context attention a major prefill bottleneck. 
Block sparse attention reduces computation by skipping unimportant blocks of the attention matrix, but scoring blocks accurately can itself be expensive.
We propose Delta-Based Tile Scoring (DTS), a training-free, content-adaptive tile scoring method for sparse block attention. DTS selects query and key anchors according to local vector changes, using fewer representatives in redundant regions while retaining more detail in heterogeneous regions. 
To accelerate anchor selection, parallel delta packing partitions the query and key sequences into tile-aligned chunks and performs bounded local scans over these chunks concurrently.
The selected tiles are then computed using the original query, key, and value vectors.
On LLaMA-3.1-8B-Instruct, DTS keeps average RULER and LongBench scores within 0.10 percentage points of full attention, with RULER evaluated up to 128K tokens. At 128K tokens, DTS reduces measured scoring-pipeline latency by 38\% compared with XAttention. On four NVIDIA H100 GPUs, it achieves a 3.3$\times$ prefill speedup over a PyTorch full-attention baseline using FlashAttention-2 at 512K tokens.

\end{abstract}


\section{Introduction}

Large language models (LLMs) increasingly process long documents, retrieved passages, and codebases~\cite {dubey2024llama3, qwen25, Qwen2-VL, reid2024gemini, liu2023ring, yang2024sweagent, lewis2020retrieval}. 
These workloads can produce long input prompts, increasing the time required before a model generates its first response token.
During the prefill stage, the model processes the input prompt and computes attention across its tokens. The number of query--key interactions grows quadratically with the input length, making attention a major source of prefill latency. 

\begin{figure}[t]
\centering
\includegraphics[width=1\columnwidth]{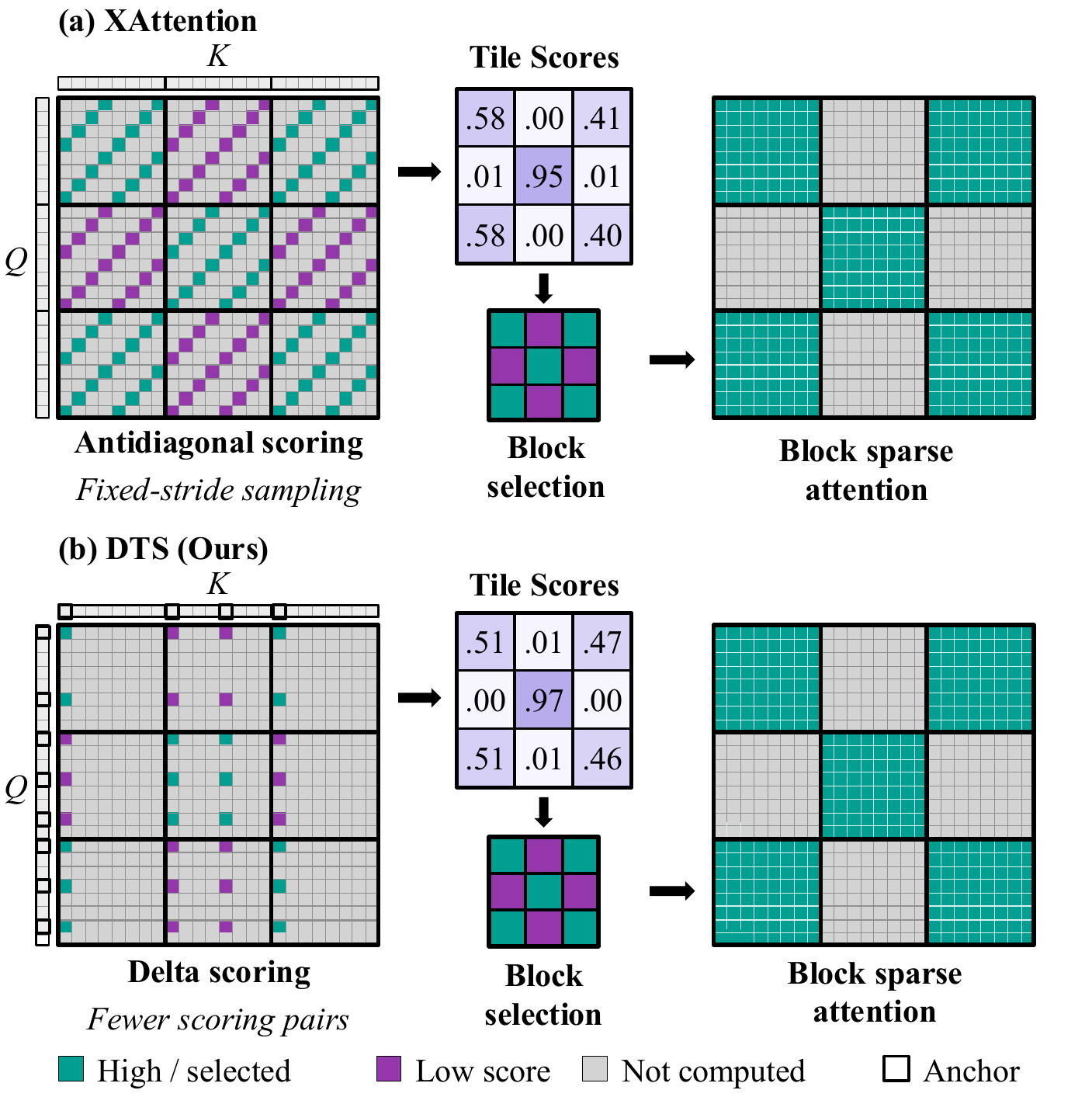}
\caption{Comparison of XAttention~\cite{xu2025xattention} and DTS. (a) XAttention samples strided antidiagonals at a fixed density. (b) DTS selects query and key anchors based on local vector changes, reducing redundant scoring in similar regions.}
\label{fig:1}
\end{figure}

Efficient dense attention kernels address part of this problem. FlashAttention~\cite{dao2022flashattention,dao2023flashattention2,shah2024flashattention3} reduces frequent memory traffic and avoids storing the full attention matrix, but still computes all query--key interactions. Its arithmetic cost therefore remains quadratic.
In fact, attention matrices are sparse: most tokens attend strongly to only a small fraction of others.
Sparse attention seeks further savings by skipping key--query interactions that contribute little to the attention output.
Early approaches~\cite{child2019generating, Beltagy2020Longformer,zaheer2020bigbird} observed that trained Transformer attention matrices concentrate on local windows and a few distant positions.
They proposed using fixed sparse patterns (local windows combined with strided or fixed summary positions) to approximate full attention with fewer operations.
The limitation is that a fixed pattern cannot adapt to the content of a given input. This motivated a shift to dynamic sparse attention, where the sparsity pattern is determined on the current input. 
For efficient execution on GPUs, block-sparse methods group interactions into blocks, or tiles, and compute the selected blocks using dense operations.

Recent dynamic sparse attention methods exploit recurring attention patterns to guide sparse computation.
MInference~\cite{jiang2024minference} observes that LLM attention heads follow consistent structural patterns (A-shape, Vertical-Slash, and Block-Sparse) across inputs. It assigns each head its dominant pattern offline and builds sparse indices accordingly at inference time.
FlexPrefill~\cite{lai2025flexprefill} selects a head's attention pattern online. It compares a pooled estimate with a reference attention distribution for a subset of queries using Jensen--Shannon distance. When the estimate is sufficiently close, it uses query-aware block selection; otherwise, it uses a Vertical-Slash pattern.
For block importance estimation, MInference’s block-sparse path and FlexPrefill’s query-aware path both reduce each query or key block to a single average vector. Such pooling is cheap, but it collapses within-block variation into one vector and can obscure the contribution of a few important token interactions.

SpargeAttn~\cite{zhang2025spargeattn} mitigates the risk by combining such pooled representatives with a check of similarity within each query and key block. Tiles involving a query or key block that fail the similarity check are retained at this stage.
This protects potentially important interactions, but reduces the number of tiles that the initial predictor can discard.
XAttention~\cite{xu2025xattention} (Fig.~\ref{fig:1}(a)) takes a different approach: it estimates block importance by summing query–key scores sampled along strided antidiagonals within each block. It selects high-scoring blocks until their cumulative normalized score reaches a specified threshold, then computes attention only on the selected blocks. However, at a given stride, its sampling density does not adapt to local differences. Regions containing many similar vectors still incur the predefined sampling cost. Increasing the stride reduces this cost but can miss important attention patterns. 
These trade-offs motivate a scoring representation that uses fewer vectors in redundant regions while retaining more detail where the vectors change.

To address these limitations, we propose \textbf{Delta-based Tile Scoring (DTS)} to reduce block importance estimation cost while maintaining inference accuracy. 
As shown in Fig.~\ref{fig:1}(b), building on the local similarity of query~\cite{wang2024kvmerger, qi2024deltallm} and key vectors,
DTS selects representative vectors, called anchors. 
Within each chunk, it starts with the first vector and selects a new anchor when the current vector differs sufficiently from the last selected anchor. Similar regions therefore use fewer scoring vectors, while blocks with greater variation can use multiple anchors. 

Our contributions are summarized as follows:
\begin{itemize}
    \item We propose a delta-based tile scoring method that leverages the similarity to select query and key representatives based on local vector changes. It uses fewer representatives in regions with similar vectors to reduce redundant scoring computations while maintaining accuracy.
    \item We design parallel delta packing with independent scans over tile-aligned chunks, limiting sequential dependencies while preserving the token positions needed to score the original tiles.
    \item We evaluate DTS on RULER and LongBench. On LLaMA-3.1-8B-Instruct, DTS with cosine threshold 0.72 keeps text benchmark averages within 0.10 points of full attention and reduces measured scoring time by 38\% relative to XAttention (stride 16) at 128K tokens.
\end{itemize}
\section{Observations}
\label{sec:observations}

We examine whether neighboring tokens produce similar query
and key vectors. We evaluate LLaMA-3.1-8B-Instruct and
Qwen2.5-7B-Instruct-1M on 50 distinct PG-19
books~\cite{rae2019compressive} and 50 non-overlapping
WikiText-103 source windows at 4K, 32K, and 128K tokens.
Each shorter input is a prefix of its 128K counterpart.
We collect query and key vectors after rotary positional
encoding (RoPE) from full-attention forward passes and compute
cosine similarity between adjacent vectors. Results average
all layers and heads, with equal weight given to each input;
duplicated key heads are counted once.

To test whether similarity is associated with local adjacency,
we randomly permute the positions of the collected vectors
and repeat the measurement. Natural order compares neighboring
positions in the original sequence, whereas shuffled order
compares neighboring entries after permutation. Both use the
same computed vectors, so the control changes their adjacency
without changing the vectors themselves.

\begin{figure}[t]
  \centering
  \includegraphics[width=0.95\columnwidth]{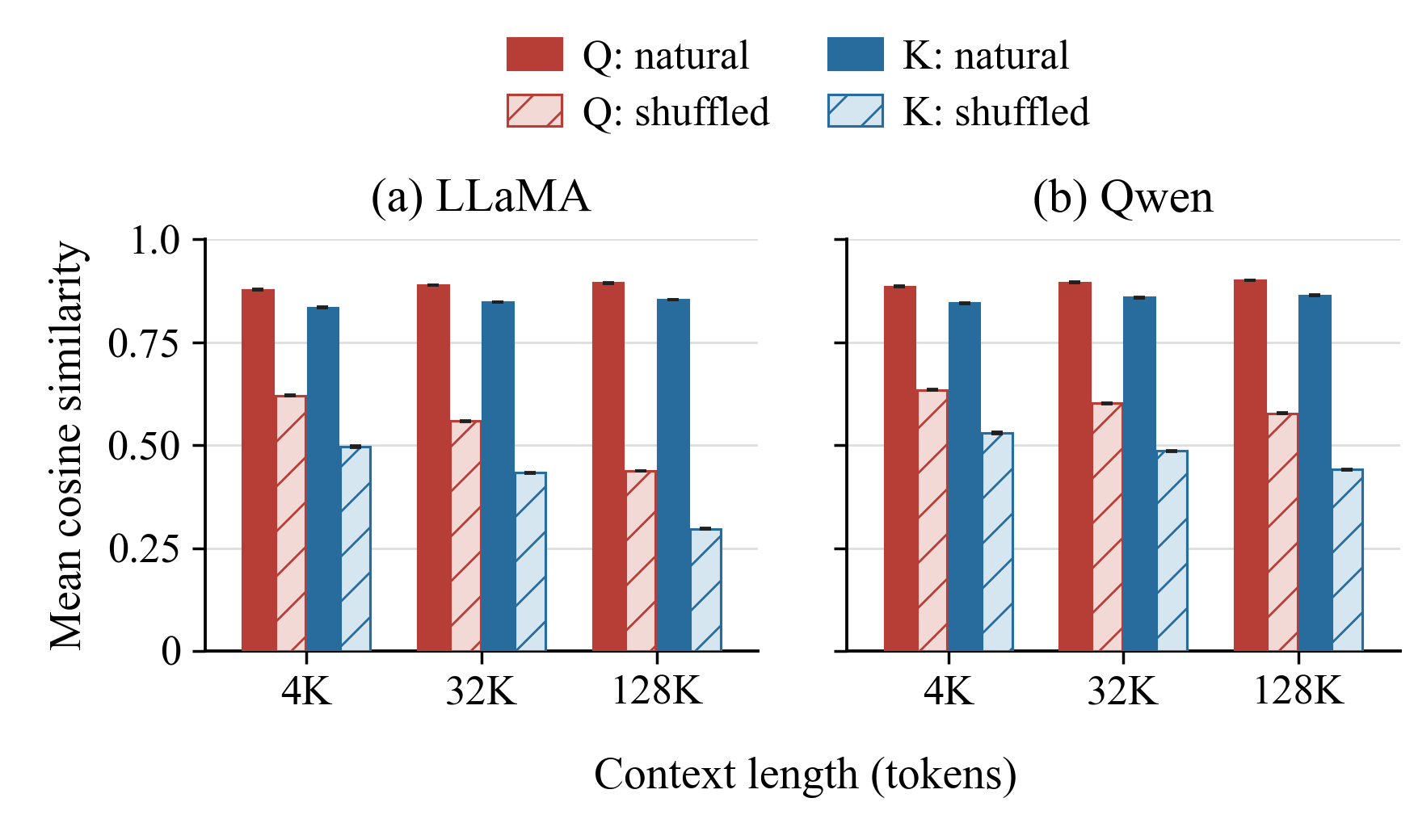}
  \caption{Mean cosine similarity of neighboring query ($Q$) and
  key ($K$) vectors in natural and shuffled order across context
  lengths. Results average all layers and heads over 50 PG-19
  books and 50 WikiText-103 windows per model. 
  }
  \label{fig:combined_similarity}
\end{figure}

As shown in Fig.~\ref{fig:combined_similarity}, natural-order
mean cosine similarity ranges from 0.878 to 0.900 for queries
and from 0.835 to 0.864 for keys across the two models and
three context lengths. Both consistently exceed their
shuffled-order counterparts; the same pattern holds when
the two corpora are analyzed separately. These results show
that local similarity is present in both queries and keys
across the evaluated settings. They suggest an opportunity
to estimate tile importance using fewer vectors from locally
similar regions, reducing repeated scoring work.

\section{Methodology}

\begin{figure}[t]
\centering
\includegraphics[width=1\columnwidth]{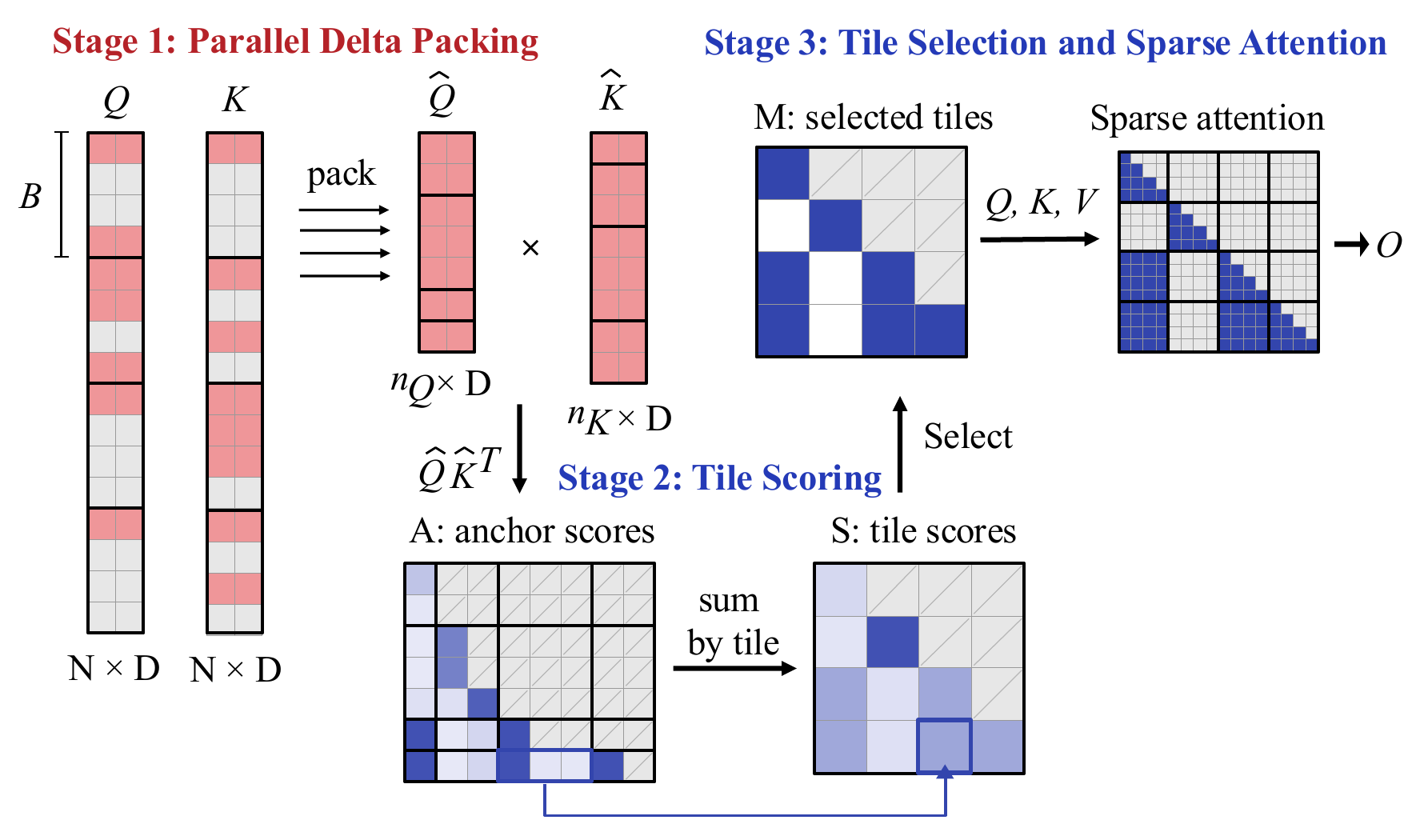}
\caption{DTS workflow. 
Parallel delta packing selects query and key anchors. Their causal scores are summed by original tile indices to guide tile selection. Sparse attention uses the original $Q,K,V$.
}
\label{fig:3}
\end{figure}

This section demonstrates three stages of DTS: 1) parallel delta packing, 2) delta-based tile scoring, and 3) tile selection and sparse attention (Fig.~\ref{fig:3} and Algorithm~\ref{alg:dts}). Per-head tile-selection thresholds are calibrated offline.

\subsection{Parallel Delta Packing}

For one attention head, let $Q,K,V\in\mathbb{R}^{N\times D}$, where $N$ is the sequence length and $D$ is the head dimension. Stage~1 applies delta packing independently to $Q$ and $K$. For either matrix $X\in\{Q,K\}$, we divide the sequence into chunks of $B$ tokens, matching the boundaries of the $B\times B$ attention tiles. Each chunk starts with its first vector as an anchor and is processed independently.

Within each chunk, we compare each vector $x_i$ with the most recently selected anchor $x_a$. A new anchor is selected when
\begin{equation}
    \cos(x_i,x_a)<\tau,
    \label{eq:delta_anchor_rule}
\end{equation}
where $\tau$ is the packing threshold. Otherwise, $x_i$ is dropped from scoring and the anchor stays unchanged. This uses fewer scoring vectors in locally similar regions and adds anchors when vectors change. Selected anchors are original vectors, without averaging. This cosine rule is used in text experiments.

Anchor updates are sequential within a chunk, while different chunks are scanned in parallel. 
Each scan processes at most $B$ vectors, while the total comparison work across all chunks remains $O(ND)$.
After anchor selection, the binary mask $\Delta$ marks selected
vectors with $\Delta[i]=1$. A prefix sum counts the selected vectors
up to each position and assigns each selected vector a packed
position $p[i]=\sum_{j=0}^{i}\Delta[j]-1$.
We then copy the selected vectors into $\hat X$ at these positions
in parallel and record their original token indices in $t$.
Algorithm~\ref{alg:delta_pack} gives the cosine variant, applied
separately to $Q$ and $K$ to obtain $(\hat Q,t^Q)$ and $(\hat K,t^K)$.

\begin{algorithm}[t]
\caption{Parallel Delta Packing}
\label{alg:delta_pack}
\footnotesize
\begin{algorithmic}[1]
\Require $X\in\mathbb{R}^{N\times D}$, threshold $\tau$,
         chunk size $B$
\Ensure Packed vectors $\hat X$ and original token indices $t$

\Procedure{DeltaPack}{$X,\tau,B$}
    \State $\Delta\gets\mathbf{0}\in\{0,1\}^{N}$
        \Comment{Anchor flags}

    \For{$c=0,\ldots,\lceil N/B\rceil-1$ \textbf{in parallel}}
        \State $s\gets cB$; $e\gets\min(s+B,N)$
        \State $a\gets s$; $\Delta[s]\gets1$
            \Comment{First anchor}
        \For{$i=s+1,\ldots,e-1$}
            \If{$\cos(X[i],X[a])<\tau$}
                \State $\Delta[i]\gets1$; $a\gets i$
                    \Comment{New anchor}
            \EndIf
        \EndFor
    \EndFor

    \State $p\gets\operatorname{cumsum}(\Delta)-1$
        \Comment{Packed positions}
    \State $n\gets\sum_i\Delta[i]$
        \Comment{Anchor count}
    \State Allocate $\hat X\in\mathbb{R}^{n\times D}$
           and $t\in\mathbb{N}^{n}$

    \For{each $i$ with $\Delta[i]=1$ \textbf{in parallel}}
        \State $\hat X[p[i]]\gets X[i]$
        \State $t[p[i]]\gets i$
            \Comment{Original position}
    \EndFor

    \State \Return $\hat X,t$
\EndProcedure
\end{algorithmic}
\end{algorithm}

\subsection{Delta-Based Tile Scoring and Selection}

\label{sec:delta_scoring}

DTS scores interactions between the $n_Q$ query anchors and $n_K$ key anchors (Stage 2). We compute scaled dot products from $\hat Q\hat K^{\top}$, exponentiate them, and apply a causal mask to obtain anchor-pair scores:
\begin{equation}
    A_{uv}=\begin{cases}
    \exp\!\left(\hat q_u^{\top}\hat k_v/\sqrt D\right),
       & t^K_v\le t^Q_u,\\
    0, & \text{otherwise}.
    \end{cases}
    \label{eq:anchor_scores}
\end{equation}
Here $u$ and $v$ index rows of $\hat Q$ and $\hat K$.

Let $T=\lceil N/B\rceil$ be the number of blocks along each axis.
The packed anchor indices belonging to query block $a$ and key block $b$
are
\begin{equation}
\begin{aligned}
    \mathcal U_a&=\{u:\lfloor t^Q_u/B\rfloor=a\},\\
    \mathcal V_b&=\{v:\lfloor t^K_v/B\rfloor=b\}.
\end{aligned}
\end{equation}
We sum their pair scores to obtain the tile importance matrix $S\in\mathbb{R}^{T\times T}$:
\begin{equation}
    S_{ab}=\sum_{u\in\mathcal U_a}\sum_{v\in\mathcal V_b}A_{uv}.
    \label{eq:tile_scores}
\end{equation}
Thus, only selected anchor pairs contribute to tile scoring. 
We use an unnormalized sum so that the tile score reflects both anchor-pair affinity and local variation within the query and key blocks. Blocks with greater variation produce more anchors and therefore receive higher selection priority.
Stage 3 normalizes each causal tile row into estimated tile weights,
\begin{equation}
    w_{ab}=\frac{S_{ab}}{\sum_{j=0}^{a}S_{aj}},\qquad 0\le b\le a.
    \label{eq:tile_weights}
\end{equation}
We sort these weights in descending order and select the shortest prefix whose cumulative weight reaches $\theta_h$. Here $h$ identifies a head within a layer. We then add the first and diagonal tiles to form the binary mask $M$. Sparse attention uses $M$ and the original $Q,K,V$, with token-level causal masking inside selected tiles. Algorithm~\ref{alg:dts} combines all three stages.

\begin{algorithm}[t]
\caption{DTS for Sparse Block Attention}
\label{alg:dts}
\footnotesize
\begin{algorithmic}[1]
\Require $Q,K,V\in\mathbb{R}^{N\times D}$, tile size $B$,
    packing threshold $\tau$, selection threshold $\theta_h\in(0,1]$
\Ensure Attention output $O$

\State $T\gets\lceil N/B\rceil$
\Statex \textit{Stage 1: Parallel Delta Packing}
\State $(\hat Q,t^Q)\gets\Call{DeltaPack}{Q,\tau,B}$
\State $(\hat K,t^K)\gets\Call{DeltaPack}{K,\tau,B}$

\Statex \textit{Stage 2: Tile Scoring}
\State $S\gets\mathbf{0}\in\mathbb{R}^{T\times T}$
\For{each tile $(a,b)$, $0\le b\le a<T$, \textbf{in parallel}}
    \State $\mathcal U_a\gets\{u:\lfloor t^Q_u/B\rfloor=a\}$
        \Comment{Query anchors}
    \State $\mathcal V_b\gets\{v:\lfloor t^K_v/B\rfloor=b\}$
        \Comment{Key anchors}
    \State $S_{ab}\gets\operatorname{ReduceSum}\limits_{\substack{
        u\in\mathcal U_a,\ v\in\mathcal V_b\\t^K_v\le t^Q_u}}
        \exp\!\left(\hat q_u^{\top}\hat k_v/\sqrt D\right)$
\EndFor

\Statex \textit{Stage 3: Tile Selection and Sparse Attention}
\State $M\gets\mathbf{0}\in\{0,1\}^{T\times T}$
\For{$a=0,\ldots,T-1$ \textbf{in parallel}}
    \State $w_{ab}\gets S_{ab}/\sum_{j=0}^{a}S_{aj}$
        for $b=0,\ldots,a$
    \State $\pi\gets$ indices $0,\ldots,a$ sorted by decreasing $w_{ab}$
    \State $m\gets0$ \Comment{Cumulative weight}
    \For{$b$ in $\pi$}
        \If{$m\ge\theta_h$}
            \State \textbf{break}
        \EndIf
        \State $M_{ab}\gets1$; $m\gets m+w_{ab}$
    \EndFor
    \State $M_{a0}\gets1$; $M_{aa}\gets1$
        \Comment{Required tiles}
\EndFor
\State $O\gets\Call{CausalBlockSparseAttention}{Q,K,V,M}$
\State \Return $O$
\end{algorithmic}
\end{algorithm}

\subsection{Per-Head Calibration}

Because $w_{ab}$ only approximates the true tile-level attention distribution, selecting tiles whose cumulative estimated weight reaches $0.9$ does not necessarily retain $90\%$ of the full-attention mass. We therefore calibrate cutoff $\theta_h$ offline for each layer and head using a target attention coverage $r$.

For calibration, we compute full-attention probabilities from
the original $Q,K$ using a causal softmax for each query.
For each tile $(a,b)$, we sum these probabilities over keys
in block $b$ and average over queries in block $a$.
The resulting reference tile weights sum to one for each
query block. Summing these weights over the selected key
blocks $b$ gives the attention coverage for query block $a$.

For each query block, we sort tiles by their reference weights and select the shortest prefix whose cumulative weight reaches $r$, forming a reference set. We then follow the DTS ordering
in Algorithm~\ref{alg:dts} and find the shortest prefix containing every tile in this reference set. The cumulative estimated weight of this prefix is the candidate cutoff for that query block.
For each layer and head, we average the candidate cutoffs over
query blocks within each document, then take the $90$th
percentile of these document averages as $\theta_h$.
We reuse XAttention's calibration set~\cite{xu2025xattention}:
$156$ documents, with $12$ from each of $13$ RULER task types
and contexts up to $80$K tokens.
For fixed $B$, $\tau$, and $r$, the model-specific threshold
table is computed once and reused for RULER and LongBench.
Experiments use $r=0.9$.
The calibrated thresholds target the desired coverage
empirically, without guaranteeing it for every query block.

\section{Experimental Setup}

\subsection{Models and Baselines}

We evaluate text workloads with LLaMA-3.1-8B-Instruct~\cite{dubey2024llama3} and Qwen2.5-7B-Instruct-1M~\cite{qwen25}. We compare DTS with full attention, MInference~\cite{jiang2024minference}, FlexPrefill~\cite{lai2025flexprefill}, and XAttention~\cite{xu2025xattention}, using the baselines' public implementations. The full-attention baseline uses PyTorch's scaled dot-product attention~\cite{paszke2019pytorch,vaswani2017attention} with the FlashAttention-2 backend~\cite{dao2023flashattention2}.
SpargeAttn is excluded because the implementation couples sparsity with quantization.

DTS uses the cosine packing rule with $\tau\in\{0.72,0.78\}$ on LLaMA and $\tau\in\{0.75,0.78\}$ on Qwen. Per-head thresholds are calibrated with target coverage $r=0.90$. XAttention uses strides $s\in\{8,16\}$ and a coverage target of $0.90$; FlexPrefill uses $\gamma=0.90$. These parameters specify each method's operating point and do not imply identical achieved coverage.

\subsection{Benchmarks and Metrics}
\label{sec:benchmarks}

RULER~\cite{hsieh2024ruler} evaluates retrieval and reasoning over long contexts. We use 13 tasks with 100 samples per task at each of six context lengths: 4K, 8K, 16K, 32K, 64K, and 128K. We report task-averaged accuracy at each length and the average across lengths.

LongBench~\cite{bai2023longbench} covers question answering, summarization, few-shot learning, synthetic tasks, and code completion. Following XAttention~\cite{xu2025xattention}, our main evaluation uses 16 tasks with 100 samples per task. We report each task's benchmark score and their unweighted average. The packing-threshold sweep uses reduced sets: nine RULER tasks with 30 samples per task, and nine LongBench tasks.

\subsection{Implementation and Latency Measurement}
\label{sec:implementation_latency}

Experiments use batch size 1. DTS uses $B=128$ for both packing chunks and attention tiles. The latency experiment distributes model layers across four NVIDIA H100 GPUs.
Following the context-length latency evaluation in MInference~\cite{jiang2024minference}, we benchmark prefill independently of downstream task evaluation using sampled WikiText-103 input at lengths from 4K to 512K. 
We also measure prefill latency on 10 inputs from each of the 13 RULER tasks at 128K, giving 130 latency measurements per model for the accuracy--latency comparison.
We measure wall-clock time for the Transformer backbone forward pass, excluding the language-model output head and decoding, with the key--value cache disabled and all GPUs synchronized before and after each run. We perform 2 warm-up runs and 10 measured runs and report mean latency. 

Density is the fraction of causal tiles selected for attention. Estimated scoring ratio multiplies the query and key anchor fractions, each averaged over heads, and then averages over layers. It estimates scoring work rather than elapsed time. Score (ms) sums the profiled packing, scoring, and selection stages; Attn (ms) measures the selected-tile attention kernel.

\section{Results}
\label{sec:results}

\subsection{Accuracy on Long-Context Tasks}
\label{sec:results_accuracy}

\begin{table}[!ht]
\centering
\caption{RULER accuracy (\%) comparing delta-based tile scoring
against baseline and block-sparse methods on LLaMA-3.1 8B
Instruct and Qwen2.5-7B-Instruct-1M. Best and second-best scores are shown in \textbf{bold}
and \underline{underlined}.}
\label{tab:delta_tile_scoring}
\resizebox{\linewidth}{!}{%
\begin{tabular}{lrrrrrrr}
\toprule
Method & 4K & 8K & 16K & 32K & 64K & 128K & Avg \\
\midrule

\multicolumn{8}{l}{\textit{LLaMA-3.1 8B Instruct}} \\

Full attention
& \underline{96.16}
& 93.77
& \underline{93.43}
& 90.47
& \textbf{86.14}
& 74.46
& \textbf{89.07} \\

MInference
& 96.04
& 93.95
& 92.60
& 88.43
& 84.66
& 63.73
& 86.57 \\

FlexPrefill
& 95.37
& 92.29
& 91.23
& 88.02
& 83.56
& 74.23
& 87.45 \\

XAttn $s{=}16$
& \textbf{96.17}
& \underline{93.97}
& \underline{93.43}
& \underline{90.56}
& 83.04
& 70.96
& 88.02 \\

XAttn $s{=}8$
& 95.57
& 93.92
& 92.71
& \textbf{90.60}
& 81.83
& 71.57
& 87.70 \\

\midrule

DTS cos\,0.72 (ours)
& 95.83
& \textbf{94.22}
& 93.36
& 90.42
& \underline{85.19}
& \underline{74.82}
& 88.97 \\

DTS cos\,0.78 (ours)
& 96.14
& 93.70
& \textbf{93.44}
& 90.35
& 84.93
& \textbf{75.55}
& \underline{89.02} \\

\midrule
\multicolumn{8}{l}{\textit{Qwen2.5-7B-Instruct-1M}} \\

Full attention
& \underline{94.10}
& \underline{93.67}
& 93.54
& \textbf{91.17}
& \underline{88.07}
& \textbf{81.41}
& \underline{90.33} \\

MInference
& \textbf{94.16}
& \textbf{94.00}
& \textbf{94.03}
& 90.99
& 87.68
& 81.23
& \textbf{90.35} \\

FlexPrefill
& 94.01
& 91.67
& 91.25
& 86.37
& 83.50
& 79.07
& 87.65 \\

XAttn $s{=}16$
& 94.02
& 93.55
& \underline{93.66}
& 90.85
& \textbf{88.21}
& 81.22
& 90.25 \\

XAttn $s{=}8$
& 94.06
& 93.53
& 93.60
& \underline{91.02}
& 87.96
& 81.23
& 90.23 \\

\midrule

DTS cos\,0.75 (ours)
& 93.87
& 93.25
& 93.24
& 90.02
& 86.09
& 79.90
& 89.39 \\

DTS cos\,0.78 (ours)
& 93.98
& 93.49
& 93.33
& 89.47
& 86.95
& \underline{81.28}
& 89.75 \\

\bottomrule
\end{tabular}%
}
\end{table}

\begin{table*}[ht]
\centering
\small
\caption{LongBench scores on LLaMA-3.1 8B Instruct and
Qwen2.5-7B-Instruct-1M. NQA = NarrativeQA, Qas = Qasper,
MFQ = MultifieldQA-en, HotP = HotpotQA, 2Wiki =
2WikiMultiHopQA, Mus = MuSiQue, GovR = GovReport, QMS =
QMSum, MNews = MultiNews, TREC = TREC, TQA = TriviaQA,
SAM = SAMSum, PCnt = PassageCount, PassR =
PassageRetrieval-en, LCC = LCC, and RepoB = RepoBench-P.
Best and second-best scores are shown in \textbf{bold} and \underline{underlined}.}
\label{tab:longbench}
\resizebox{\linewidth}{!}{%
\begin{tabular}{lrrrrrrrrrrrrrrrrr}
\toprule
Method & NQA & Qas & MFQ & HotP & 2Wiki & Mus & GovR & QMS
& MNews & TREC & TQA & SAM & PCnt & PassR & LCC & RepoB & Avg \\
\midrule
\multicolumn{18}{l}{\textit{LLaMA-3.1 8B Instruct}} \\

Full attention
& \textbf{31.91}
& \underline{43.06}
& 56.25
& \underline{54.87}
& 47.97
& \textbf{34.88}
& 34.47
& 25.27
& \textbf{27.15}
& \textbf{67.00}
& 89.47
& 44.56
& \textbf{7.58}
& \textbf{99.00}
& 51.94
& 45.12
& \textbf{47.53} \\

MInference
& 31.74
& 42.25
& 56.50
& 54.70
& \textbf{50.17}
& 31.80
& 34.51
& 25.27
& \underline{27.13}
& \textbf{67.00}
& 89.69
& 44.09
& 5.00
& \underline{94.00}
& 52.14
& 47.32
& 47.08 \\

FlexPrefill
& 26.02
& 39.00
& 55.19
& \textbf{55.11}
& 38.53
& 28.73
& 33.44
& 25.53
& 26.93
& \textbf{67.00}
& 89.29
& 44.58
& 2.00
& 83.00
& \underline{54.02}
& 48.07
& 44.78 \\

XAttn $s{=}16$
& 31.01
& 37.56
& 56.65
& 53.35
& \underline{48.37}
& 32.71
& \underline{34.83}
& 24.37
& 26.46
& \textbf{67.00}
& \textbf{89.80}
& \textbf{45.42}
& 5.33
& 92.00
& 53.94
& 44.74
& 46.47 \\

XAttn $s{=}8$
& 29.84
& 41.08
& 54.43
& 53.23
& 47.59
& 31.01
& 34.49
& \textbf{26.37}
& 26.46
& \textbf{67.00}
& \underline{89.77}
& 44.53
& 4.00
& 93.00
& 52.67
& 48.61
& 46.50 \\

\midrule

DTS cos\,0.72 (ours)
& 31.62
& 42.86
& \underline{57.04}
& 53.29
& 47.88
& \underline{32.87}
& 34.67
& \underline{25.66}
& 26.97
& 64.00
& 89.30
& \underline{44.74}
& 6.50
& \textbf{99.00}
& \textbf{54.32}
& \textbf{49.48}
& \underline{47.51} \\

DTS cos\,0.78 (ours)
& \underline{31.81}
& \textbf{43.60}
& \textbf{57.19}
& 54.51
& 45.27
& 32.72
& \textbf{35.26}
& 25.35
& 26.67
& \underline{66.00}
& 89.30
& 43.83
& \underline{6.83}
& \textbf{99.00}
& 50.62
& \underline{49.26}
& 47.33 \\

\midrule
\multicolumn{18}{l}{\textit{Qwen2.5-7B-Instruct-1M}} \\

Full attention
& 29.26
& 45.12
& 51.49
& \underline{58.34}
& \underline{55.45}
& 32.70
& 35.22
& 23.99
& 25.69
& 71.00
& 83.31
& 45.24
& \underline{10.00}
& \textbf{100.00}
& 48.40
& 32.92
& 46.76 \\

MInference
& 29.38
& \underline{45.36}
& 51.85
& 56.77
& 54.65
& \textbf{33.47}
& 35.20
& 24.18
& 25.98
& 71.00
& 83.31
& \textbf{46.32}
& \underline{10.00}
& \underline{99.00}
& 47.63
& 33.44
& 46.72 \\

FlexPrefill
& 24.68
& 39.46
& 50.63
& 48.72
& 50.58
& 21.06
& 33.61
& \textbf{24.67}
& 25.42
& 70.00
& 78.64
& 46.11
& 2.00
& 59.00
& 45.18
& 33.61
& 40.84 \\

XAttn $s{=}16$
& 29.84
& 44.74
& 51.41
& 57.63
& \textbf{55.62}
& 32.60
& 34.84
& 24.00
& \textbf{26.09}
& 70.00
& 83.18
& 46.00
& \textbf{12.00}
& \textbf{100.00}
& 46.70
& 31.97
& 46.66 \\

XAttn $s{=}8$
& \underline{30.26}
& \textbf{45.77}
& 52.31
& 58.16
& 53.78
& 31.62
& 35.18
& 24.09
& \underline{26.00}
& 71.00
& 83.51
& 45.83
& 9.00
& \textbf{100.00}
& 45.14
& 31.70
& 46.46 \\

\midrule

DTS cos\,0.75 (ours)
& \textbf{30.98}
& 45.18
& \underline{53.00}
& \textbf{58.91}
& 51.57
& 31.73
& \textbf{35.58}
& \underline{24.57}
& 25.95
& \underline{72.00}
& \textbf{84.55}
& \underline{46.29}
& 4.00
& \textbf{100.00}
& \textbf{48.78}
& \textbf{36.42}
& \textbf{46.84} \\

DTS cos\,0.78 (ours)
& 29.10
& 44.66
& \textbf{53.08}
& 56.61
& 54.63
& \underline{33.21}
& \underline{35.47}
& 24.12
& 25.64
& \textbf{74.00}
& \underline{84.19}
& 46.13
& 6.00
& \textbf{100.00}
& \underline{48.49}
& \underline{33.71}
& \underline{46.82} \\

\bottomrule
\end{tabular}%
}
\end{table*}

\paragraph{RULER}
On LLaMA, DTS at $\tau=0.78$ achieves an average score of 89.02, within 0.05 points of full attention (89.07) and 1.00--2.45 points above all baseline sparse methods (Table~\ref{tab:delta_tile_scoring}). At 128K, DTS attains the highest score of 75.55, exceeding full attention by 1.09 points and XAttention by 3.98--4.59 points. On Qwen, DTS at $\tau=0.78$ averages 89.75, remaining within 0.60 points of full attention and the strongest sparse baselines. At 128K, it achieves 81.28, only 0.13 points below full attention and the highest score among the sparse methods, slightly exceeding MInference (81.23) and XAttention (81.22--81.23).

\paragraph{LongBench}
On LLaMA, DTS at $\tau=0.72$ achieves an average score of 47.51, within 0.02 points of full attention and 1.01--1.04 points above XAttention (Table~\ref{tab:longbench}). It achieves the best scores on both code tasks, LCC and RepoBench-P, and matches full attention on PassageRetrieval-en; the $\tau=0.78$ setting additionally leads on Qasper, MultifieldQA-en, and GovReport. On Qwen, DTS at $\tau=0.75$ obtains the highest average score of 46.84, exceeding full attention by 0.08 points, MInference by 0.12 points, and XAttention by 0.18--0.38 points, while $\tau=0.78$ remains competitive at 46.82. Across the two operating points, DTS achieves the best score on eight of the 16 tasks and ties for the best on PassageRetrieval-en, demonstrating strong performance across question answering, retrieval, and code tasks.

\subsection{Prefill Latency and Scoring Cost}
\label{sec:results_latency}
\begin{figure*}[t]
\centering
\includegraphics[width=2\columnwidth]{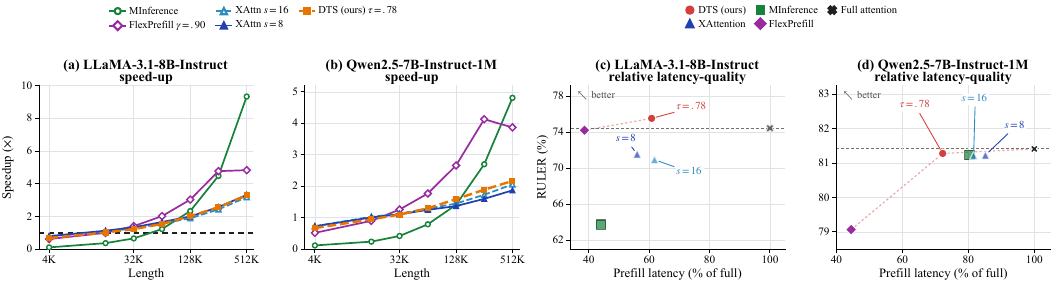}
\caption{
Prefill efficiency and accuracy. (a)–(b) Prefill speedup relative to full attention across context lengths, measured using WikiText-103. (c)–(d) RULER accuracy versus prefill latency at 128K, measured on RULER, for LLaMA and Qwen, 100\% corresponds to 17.33\,s and 13.82\,s, respectively.
}
\label{fig:speedup}
\end{figure*}

Fig.~\ref{fig:speedup}(a) and (b) show how prefill speedup
changes with sequence length. DTS becomes beneficial as the context grows and reaches $3.29\times$ on LLaMA and $2.17\times$ on Qwen at 512K. At 128K, DTS achieves $2.02\times$ and $1.59\times$, respectively, compared with $1.89\times$/$2.00\times$ for XAttention on LLaMA and
$1.45\times$/$1.37\times$ on Qwen for $s=16$/$s=8$. MInference and FlexPrefill can be faster at the longest lengths, so speed alone does not capture the quality trade-off.

Fig.~\ref{fig:speedup}(c) and (d) therefore combine the 128K
RULER scores with latency measured on RULER inputs. Latency is shown as a percentage of full attention to make the reduction comparable across models. On LLaMA, DTS attains the highest score, 75.55, while using 60.8\% of full-attention latency. It is both faster and 4.59
points more accurate than XAttention with $s=16$; XAttention with $s=8$ and FlexPrefill are faster, but score 3.98 and 1.32 points lower, respectively. On Qwen, DTS reaches 81.28 at 72.2\% of full-attention latency. At the evaluated operating points, it is faster than MInference and both XAttention settings while giving a slightly higher score. FlexPrefill reduces latency further but lowers the score to 79.07. These results place DTS on the measured Pareto frontier and show its advantage in the near-full-accuracy regime.

\begin{table}[t]
  \centering
  \caption{Scoring-pipeline cost comparison between XAttention ($s=16$) and DTS ($\tau=0.72$) on LLaMA-3.1-8B-Instruct. The scoring pipeline includes all stages required to construct the sparse tile mask and excludes block-sparse attention. Times are summed over all 32 layers.}
  \label{tab:scoring_cost}
\small
\setlength{\tabcolsep}{2.6pt}
\renewcommand{\arraystretch}{0.8}
\begin{tabular*}{\linewidth}{@{\extracolsep{\fill}}lrrrrrr@{}}
  \toprule
  & \multicolumn{2}{c}{Density (\%)}
  & \multicolumn{2}{c}{Est. score (\%)}
  & \multicolumn{2}{c}{Scoring pipeline (ms)} \\
  \cmidrule(lr){2-3}\cmidrule(lr){4-5}\cmidrule(l){6-7}
  Context & XAttn & DTS & XAttn & DTS & XAttn & DTS \\
  \midrule
  4K & 66.1 & 55.5 & 6.25 & 4.41 & 19 & 45 \\
  16K & 36.4 & 31.3 & 6.25 & 3.05 & 37 & 92 \\
  32K & 27.6 & 24.9 & 6.25 & 2.77 & 141 & 206 \\
  64K & 20.5 & 19.9 & 6.25 & 2.61 & 542 & 541 \\
  128K & 17.1 & 17.6 & 6.25 & 2.26 & 2{,}571 & 1{,}592 \\
  256K & 15.4 & 17.0 & 6.25 & 1.65 & 8{,}018 & 4{,}786 \\
  512K & 11.4 & 15.2 & 6.25 & 0.84 & 28{,}270 & 14{,}407 \\
  \bottomrule
\end{tabular*}
\end{table}

Table~\ref{tab:scoring_cost} compares the recorded scoring time of DTS and XAttention from 4K to 512K tokens. 
The measured scoring-pipeline time includes packing, scoring, and selection.
We use $\tau=0.72$ on LLaMA because it yields a tile density closely matched to XAttention with $s=16$, enabling a controlled scoring-cost comparison.
DTS has a lower estimated scoring ratio at every length,
but is slower through 32K and has nearly the same recorded scoring
time at 64K. At 128K, the recorded time decreases from 2,571 to
1,592\,ms, a 38.1\% reduction, at similar tile densities
(17.6\% for DTS and 17.1\% for XAttention). The reductions reach
40.3\% at 256K and 49.0\% at 512K. These longer settings evaluate
latency scaling, and their tile densities differ more substantially:
at 512K, DTS selects 15.2\% of tiles, compared with 11.4\% for
XAttention.

\subsection{Ablation Studies}
\label{sec:results_cos_ablation}
\subsubsection{Effect of Packing Threshold}
We vary the cosine packing threshold $\tau\in\{0.70,0.72,0.75,0.78,0.80\}$ with calibration target $r=0.90$. Table~\ref{tab:cos_ablation} combines 128K latency measurements with task scores on the reduced evaluation sets described in Section~\ref{sec:benchmarks}.
Increasing $\tau$ selects more anchors for scoring. From $\tau=0.70$ to 0.80, the estimated scoring ratio increases from 1.6\% to 8.7\%, and measured scoring time rises from 1,447 to 2,885 ms. Meanwhile, selected tile density decreases from 19.0\% to 13.3\%, reducing sparse-attention time from 3,094 to 2,140 ms. These opposing costs limit the change in prefill speedup, which ranges from 1.97$\times$ to 2.10$\times$. 
RULER scores stay within 0.28 points of full attention throughout this sweep. Among the tested settings, $\tau=0.72$ gives both the highest speedup (2.10$\times$) and LongBench score (39.21), while $\tau=0.78$ gives the highest RULER score (91.80) at 2.04$\times$. These two settings illustrate the balance between scoring cost and task performance.

\begin{table}[t]
\centering
\setlength{\abovecaptionskip}{2pt}
\setlength{\belowcaptionskip}{2pt}
\caption{Cosine-threshold sweep on LLaMA-3.1-8B-Instruct.
Latency results are measured at 128K.}
\label{tab:cos_ablation}

\small
\renewcommand{\arraystretch}{0.9}
\setlength{\tabcolsep}{2.2pt}

\resizebox{\linewidth}{!}{%
\begin{tabular}{lrrrrrrr}
\toprule
\multirow{2}{*}{$\tau$}
& \multirow{2}{*}{Speedup}
& \multirow{2}{*}{Density}
& \multirow{2}{*}{\shortstack{Est. score\\ratio (\%)}}
& \multicolumn{2}{c}{Phase time (ms)}
& \multirow{2}{*}{RULER}
& \multirow{2}{*}{LongBench} \\
\cmidrule(lr){5-6}
& & & & Scoring & Attention & & \\
\midrule
Full
& 1.00$\times$ & 100\% & -- & -- & --
& 91.52 & 39.54 \\
\specialrule{0.3pt}{1pt}{1pt}
0.70
& 2.08$\times$ & 19.0\% & 1.6\% & 1{,}447 & 3{,}094
& 91.69 & 38.95 \\
0.72
& 2.10$\times$ & 17.6\% & 2.3\% & 1{,}592 & 2{,}853
& 91.51 & 39.21 \\
0.75
& 2.09$\times$ & 15.7\% & 3.8\% & 1{,}954 & 2{,}545
& 91.42 & 39.08 \\
0.78
& 2.04$\times$ & 14.2\% & 6.3\% & 2{,}445 & 2{,}284
& 91.80 & 39.15 \\
0.80
& 1.97$\times$ & 13.3\% & 8.7\% & 2{,}885 & 2{,}140
& 91.58 & 38.61 \\
\bottomrule
\end{tabular}%
}
\vspace{-1mm}
\end{table}

\begin{table}[t]
\centering
\caption{Effect of parallel delta packing on
LLaMA-3.1-8B-Instruct at $\tau=0.72$. \emph{Packing only} measures delta packing, whereas the \emph{DTS attention pipeline} includes delta packing, tile scoring, tile selection, and block-sparse attention. The sequential and parallel implementations differ only in the chunk-scan schedule. All times cover the 32 attention layers.
Timings are profiled independently of Table \ref{tab:cos_ablation}.}
\label{tab:parallel_packing}
\footnotesize
\setlength{\tabcolsep}{2.5pt}
\renewcommand{\arraystretch}{0.9}
\begin{tabular}{@{}lrrr@{\hspace{7pt}}rrr@{}}
\toprule
& \multicolumn{3}{c}{Packing only}
& \multicolumn{3}{c}{DTS attention pipeline} \\
\cmidrule(lr){2-4}
\cmidrule(lr){5-7}
Context
& Seq. (ms) & Par. (ms)& Speedup
& Seq. (ms)& Par. (ms) & Speedup \\
\midrule
8K
& 451.7 & 24.4 & $18.53\times$
& 540.5 & 110.4 & $4.90\times$ \\

32K
& 1,782.4 & 76.4 & $23.34\times$
& 2,205.5 & 482.1 & $4.57\times$ \\

64K
& 3,555.5 & 144.2 & $24.65\times$
& 4,700.3 & 1,271.1 & $3.70\times$ \\

128K
& 7,128.5 & 295.8 & $24.10\times$
& 10,931.6 & 4,094.0 & $2.67\times$ \\
\bottomrule
\end{tabular}
\end{table}

\subsubsection{Effect of Parallel Packing}
We compare two Triton implementations that scan chunks sequentially or in parallel.
As shown in Table~\ref{tab:parallel_packing}, parallel chunk scans accelerate delta packing by $18.53$--$24.65\times$. The benefit remains substantial when tile scoring, tile selection, and block-sparse attention are included: the complete DTS attention pipeline is accelerated by $2.67$--$4.90\times$. 
At 128K tokens, packing latency decreases from $7{,}128.5$ to $295.8$\,ms, while attention-pipeline latency decreases from $10{,}931.6$ to $4{,}094.0$\,ms. 
The smaller pipeline-level speedup at longer contexts occurs because tile scoring, selection, and sparse attention account for a larger fraction of the total latency.

\subsubsection{Effect of Anchor Selection}
We replace delta-based anchors with random anchors while
matching the query/key anchor counts in every layer, head, and chunk,
as well as the selected-tile count of every query block. At 128K with $\tau=0.78$ on Qwen RULER, delta-based anchors achieve 81.28,
compared with an average of 79.22 for random anchors over three seeds.

\section{Conclusion}

DTS introduces a content-adaptive tile-scoring method that selects query and key anchors at local vector changes and parallelizes packing across tile-aligned chunks. 
DTS keeps RULER and LongBench averages within 0.10 points of full attention on LLaMA, reduces scoring-pipeline latency by 38.1\% relative to XAttention at comparable tile density at 128K, and achieves a 3.3$\times$ end-to-end prefill speedup at 512K.
These results show that adaptive anchor selection can reduce
redundant scoring while retaining a favorable accuracy--latency
trade-off for long-context prefill.

\bibliographystyle{IEEEtran}

\newpage
\bibliography{references}

\end{document}